\documentclass[conference]{IEEEtran}
\ifCLASSINFOpdf
\else
\fi
\usepackage{graphicx}
\usepackage{subfig}
\usepackage{caption}
\usepackage{multirow}
\graphicspath{ {images/} }

\begin{document}
%
\title{Partial Fingerprint Detection using\\core point location}


\author{
\IEEEauthorblockN{Wajih Ullah Baig}
\IEEEauthorblockA{Technology Directorate\\
NADRA\\
G-5, Islamabad\\
wajih.ullah@nadra.gov.pk}
\and

\IEEEauthorblockN{Adeel Ejaz}
\IEEEauthorblockA{Technology Directorate\\
NADRA\\
G-5, Islamabad\\
adeel.ejaz@nadra.gov.pk}

\and
\IEEEauthorblockN{Dr. Umar Munir}
\IEEEauthorblockA{Technology Directorate\\
NADRA\\
G-5, Islamabad\\
muhammad.umer@nadra.gov.pk}
\and
\IEEEauthorblockN{Kashif Sardar}
\IEEEauthorblockA{Technology Directorate\\
NADRA\\
G-5, Islamabad\\
kashif.sardar@nadra.gov.pk}
}


%


\maketitle

\begin{abstract}
In Biometric identification, fingerprints based identification has been the widely accepted mechanism. Automated fingerprints identification/verification techniques are widely adopted in many civilian and forensic applications. In forensic applications fingerprints are usually incomplete, broken, unclear or degraded which are known as partial fingerprints. Fingerprints identification/verification largely suffer from the problem of handling partial fingerprints. In this paper a novel and simple approach is presented for detecting partial fingerprints using core point location. Our techniques is particularly useful during the acquisition stage as to determine whether a user needs to re-align the finger to ensure a complete capture of fingerprint area.This technique is tested on FVC-2002 DB1A. The results are very accurate which are presented in the "Results" sections.
\\
\end{abstract}


%
\IEEEpeerreviewmaketitle
The paper is organized into the following sections. Section \ref{Background} is an overview of AFIS, section \ref{Introduction} is an introduction to bio-metrics, section \ref{Proposed Detection Technique} describes the proposed detection technique. Section \ref{Results} is performance analysis and results. The last \ref{Conclusion} provides the conclusions.

\section{Background} \label{Background}
Automatic fingerprint identification/recognition systems (AFIS/AFRS) have been nowadays widely used in personal identification applications such as access control \cite{Automatic Fingerprint Recognition Springer,Handbook of Fingerprint Recognition}. Minutiae based fingerprints recognition techniques are widely used in those system \cite{Handbook of Fingerprint Recognition}. Minutiae based matching techniques are best suited for good quality and complete images. Improper capturing/scanning results Partial or incomplete fingerprints with poor quality \cite{SIFT Features}. General minutiae based matching algorithm does not perform well case of partial fingerprints which results in degradation of over all automatic fingerprint identification/recognition systems. In order to minimize the effect of partial fingerprints our proposed techniques detect the partial fingerprints at the time of capturing or scanning.

\section{Introduction} \label{Introduction}
Fingerprint identification is a pattern recognition problem which has been under research for many years. Achievement of very high accuracy matching with light-weight algorithms in poor quality or partial images is still an open issue. There has been a lot of effort in providing different features for matching fingerprints. Mostly minutiae based matching algorithms. Though there are efforts for matching using frequency, orientation, texture etc. In minutiae based matching, minutiae play a vital role in fingerprint comparisons.
Such algorithms solely base their performances on fingerprint quality. Core point detection can be achieved using the direction fields that are deduced from a fingerprint. For core points, the field become discontinuous or the point at which maximum direction change. Orientation field is used to determine the reference point. A filter is applied to the orientation filed to detect the maximum direction change in ridge flow. 

Human Fingerprints are unique for each individual. This uniqueness provides the ground for identification and verification of individuals and the automation of the process is termed as Automated Fingerprint Identification System (AFIS). Fingerprints are matched by encoding each fingerprint image into a numerically equivalent identity. These numerical identities are cross-matched with others to either verify or identify an individual. The numerical identities are conventionally based on minutiae. Algorithms for fingerprint recognition are mainly based on minutiae information. However, the small number of minutiae in partial fingerprints is still a challenge in fingerprint matching. For this purpose it needs to be ensured that fingerprints with acceptable quality are available. Apart from quality, fingerprints also have classified patterns called deltas,,whorls, loops, cores etc. A fingerprint matching algorithms needs to correctly perform on full fingerprints and partial fingerprints to attain high accuracy discrimination \cite{IEEEhowto:PMTRBM}. If fingerprints are known to be partial in advance, a fingerprint algorithm can be tuned to perform accordingly \cite{IEEEhowto:BioRec}.  


\section{Proposed Detection Technique} \label{Proposed Detection Technique}
\subsubsection{Pre-processing}
Prior to the extraction of core points, a fingerprint image undergoes the steps of pre-processing.
The pre-processing steps in a typical manner would follow;

\begin{enumerate}
  \item Segmentation 
  \item Basic Enhancement (Contrast stretch)
  \item Orientation field estimation
  \item Advanced Enhancement (Gabor Filtering)
  \item Binarization
  \item Orientation field estimation
\end{enumerate}

A fingerprint image is first segmented. Then passed through contrast normalization to ensure the ridges are more visible. Followed by gradient-based orientation field estimation \cite{IEEEhowto:GBOFE}. The orientation field deduced from this step is used to align the Gabor filters in the next step. That is, a Gabor filters will be aligned to the local orientation provided by the local window over which Gabor responses are calculated. After Gabor filtering a enhanced version of the  fingerprint is produced. The enahced version is binarized. Once again the orientation field is estimated from the binary image. This time a more smooth orientation field shall be available due to the fact that Gabor filtering has produced fine ridges. The  orientation field is then passed onto complex filtering for core point detection. It is interesting to note that complex filtering can also extract minutiae \cite{IEEEhowto:CompFM}. 
\subsubsection{Complex Filtering}
Core point (singular point) localization via complex filtering is a very effective technique \cite{IEEEhowto:CompF},\cite{IEEEhowto:Bigun2001} . Using multi-scale filters, the output from this technique is robust and reliable. Morphological techniques rely lesser on statistics and more on shape, where as this technique relies heavily on the statistics of the image in question. Core point is the central point of a fingerprint image. Fingerprints tend to have may have no or multiple core points. There is also classification of core points which tend to have special patterns. Some of these patterns being whorl, loops etc. In whorl patterns, the core point lies in the middle of the spiral. While in loop pattern core point lie in the in the top region of innermost loop. These patterns are easily mapped using complex filters image. Variance image is calculated and finally highest response from the variance image is used to detect a singularity if a threshold value is fulfilled.
\\\\
Raidial symmetries can be modeled using complex filters of order \textit{m} \cite{IEEEhowto:Bigun97}. These symmetries can be expressed in cartesean and polar forms by equation (\ref{eq:BasicZCartSym}) and equation (\ref{eq:BasicZPolarSym}) respectively; 
\begin{equation}\label{eq:BasicZCartSym}  z = (x+iy)  \end{equation}

\begin{equation}\label{eq:BasicZPolarSym}  z = r(cos\varphi+isin\varphi)  \end{equation}

\begin{equation}\label{eq:BasicPolarSym}  (x+iy) = re^{i\varphi} \end{equation}
In the perspective of applying the complex filters to the image, the filters have to be orientation isotropic (rotation invariant) and separable. The requirement of orientation isotropicity and separation is fulfilled by a Gaussian distribution. The Gaussian restricted by a window of size W, inhibits the properties of orientation isotropicity and separability in polar coordinates as it is a function of radius only \cite{IEEEhowto:Rudin87}. 
\\

The complex filter of order \(m\) in terms of a Gaussian filter can be represented by \(F_{gm}\)

\begin{equation}\label{cartesianMgauss} F_{gm} = ({x+iy})^m
g(x,y) \end{equation}
\begin{equation}\label{polargauss} g(x,y) =  e^{-(x^2+y^2)/2\sigma^2}
\end{equation}
\begin{equation}\label{PolarMgauss} F_{gm} =  re^{im\varphi}e^{-(x^2+y^2)/2\sigma^2}
\end{equation}

Equation (\ref{polargauss}) is the Gaussian distribution.
In order to apply the complex filter to detect singularities, 
the complex filter \(F_{gm}\) of order \(m\) is convolved with absolute valued \(\|c\|\) of complex tensor field image \(c\) instead of the original image.

\begin{equation} c =  (f_x+if_y )
\end{equation}
\begin{equation} \|c\| =  \|f_x+if_y \|
\end{equation}

\begin{equation}\label{convFgmAbsC} R =   \|c\| * F_{gm} 
\end{equation}

\begin{figure}[h]
\centering
\subfloat[Complex image]{\includegraphics[height=1.1in]{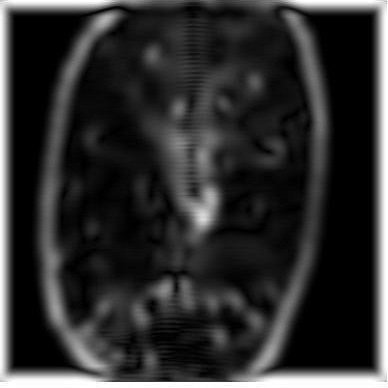}\label{fig:complex_filtered_img}}
\thinspace\thinspace
\subfloat[Variance image]{\includegraphics[height=1.1in]{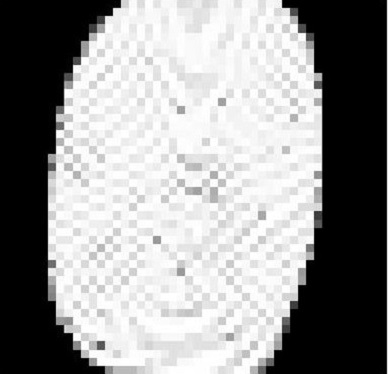}\label{fig:variance_img}}
\\
\subfloat[Core point detected]{\includegraphics[height=1.1in]{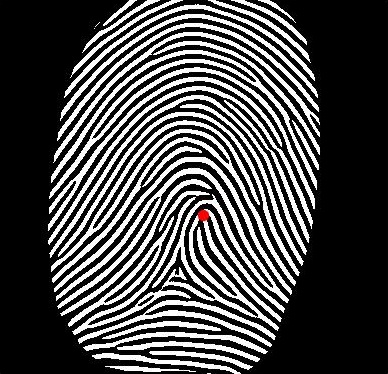}\label{fig:core_img}}
\thinspace\thinspace
\subfloat[Segmentation mask]{\includegraphics[height=1.1in]{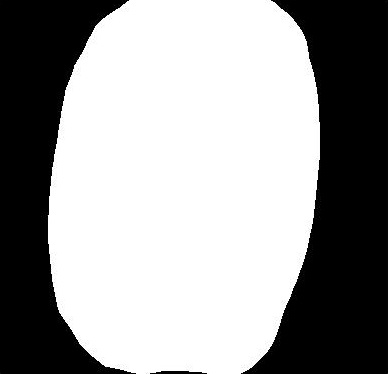}\label{fig:seg_img}}

\caption{A stepwise output presentation for complex filtering}
\end{figure}

Equations (\ref{convFgmAbsC}) represents the convolution operation of filter \(F_{gm}\) with abosolute value of \(c\). A visualization of this step can be seen in Fig \ref{fig:complex_filtered_img}. The variable \textit{f\textsubscript{x}} is the derivative of the image in x-direction and the variable \textit{f\textsubscript{y}} is the derivative in y-direction. 
\\\\
In order to detect singularities, filters of first order \(m=1\) and \(m=-1\) are used to detect singularities of type cores and deltas.

\begin{figure}[h]
\includegraphics[scale=1.1]{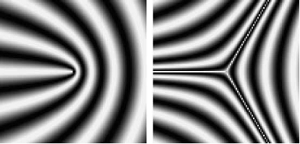}
\caption{Filters of order \(m=1\) and \(m=-1\). Singualr point detetors of type Whorls/Loops and Delta Points respectively. \cite{IEEEhowto:CompF}}
\centering
\end{figure}

\begin{figure}[h]
\includegraphics[scale=1.1]{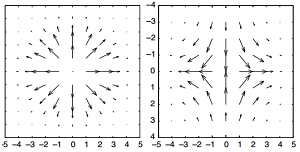}
\caption{Tensor field views of core and delta points.\cite{IEEEhowto:CompF}}
\centering
\end{figure}

\begin{figure}[h]
\centering
\subfloat[Original image]{\includegraphics[height=1.1in]{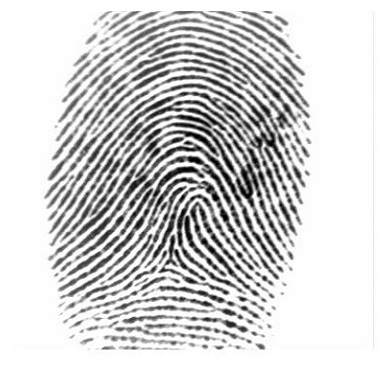}}
\thinspace\thinspace
\subfloat[Gabor filtered]{\includegraphics[height=1.1in]{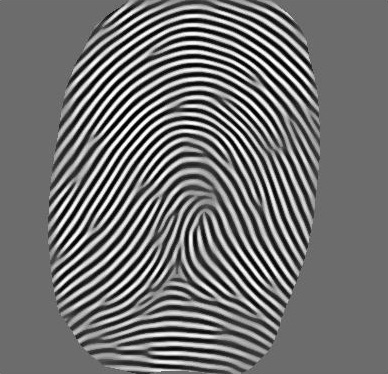}}
\\
\subfloat[Gradient x]{\includegraphics[height=1.1in]{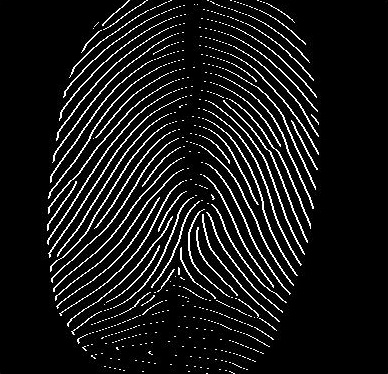}} 
\thinspace\thinspace
\subfloat[Gradient y]{\includegraphics[height=1.1in]{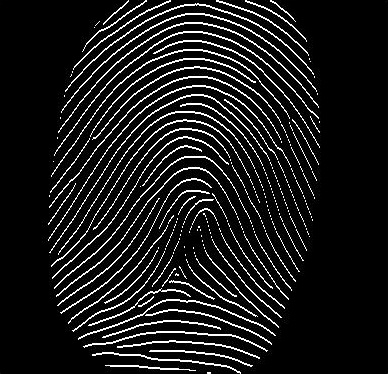}}
\\
\subfloat[Binary image]{\includegraphics[height=1.1in]{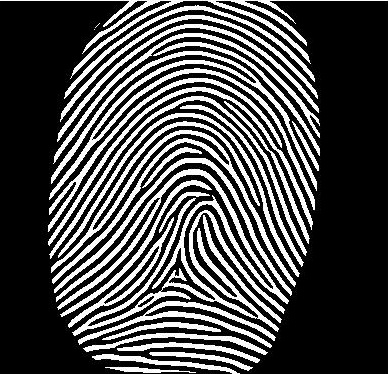}} 
\thinspace\thinspace
\subfloat[Orientation field]{\includegraphics[height=1.1in]{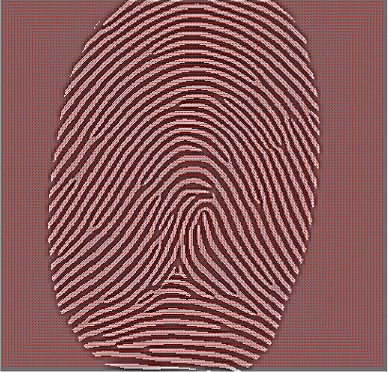}}
\\
\caption{A stepwise output presentation for complex filtering}
\end{figure}

The outcome from equation (\ref{convFgmAbsC}) is used to create a variance image Fig.\ref{fig:variance_img}. Variance is calculated in a block-wise manner in non-overlapping window that is 1/4\textsuperscript{th}  the size of the Gaussian window \(\big[ W , W \big]\). Here we set \(W=32\). The response from each block \(\big[ W/4 , W/4 \big]\) is calculated to create variance image. The image is then normalized using  the following equation.
\begin{equation}\label{normalization} I_n = (I-min(I))/(max(I)-min(I))
\end{equation}
Equation (\ref{normalization}) shall limit the variance inside the close domain of \([0,1]\).
The Singular point shall be located at the highest variance point inside the variance image, see Fig.(\ref{fig:core_img}). In the case of detecting singular points at Whorls/Loops, the order of complex filters is set \(m=1\). 

\subsubsection{Partial Image Detection}
Placing the core point as the center of the origin of the segmentation mask image (Fig.\ref{fig:seg_img}), white pixels are counted along the horizontal and vertical axes of the Cartesian coordinates whose center is the core point. The counting along the axes continues until a black pixel or image ending is encountered.

\begin{equation}\label{eq:LC} 
L_c=\sum\limits_{i=C_X}^{B_{X1}} SegmentationMask(i,Y)
\end{equation}
\begin{equation} \label{eq:RC}
R_c=\sum\limits_{i=C_X}^{B_{X2}}  SegmentationMask(i,Y)
\end{equation}
\begin{equation} \label{eq:UC}
U_c=\sum\limits_{i=C_Y}^{B_{Y1}}  SegmentationMask(X,i)
\end{equation}
\begin{equation} \label{eq:DC}
D_c=\sum\limits_{i=C_Y}^{B_{Y2}} SegmentationMask(X,i)
\end{equation}

\begin{equation}\label{eq:MAXV} 
Max_{value} = max(L_c,R_c,U_c,D_c)
\end{equation}

\begin{equation}\label{eq:LCN} 
L_{cn}=L_c/Max_{value}
\end{equation}
\begin{equation} \label{eq:RCN}
R_{cn}=R_c/Max_{value}
\end{equation}
\begin{equation} \label{eq:UCN}
U_{cn}=U_c/Max_{value}
\end{equation}
\begin{equation} \label{eq:DCN}
D_{cn}=D_c/Max_{value}
\end{equation}

\begin{figure}[h]
\centering
\subfloat[1\_1.tif]{\includegraphics[height=1.1in]{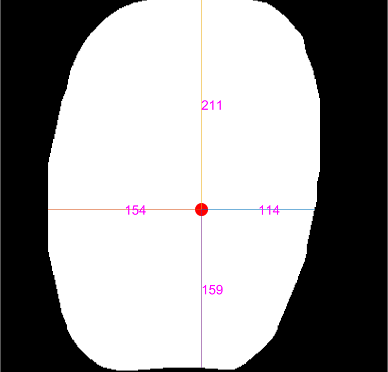}\label{fig:full_img}}
\thinspace\thinspace
\subfloat[4\_5.tif]{\includegraphics[height=1.1in]{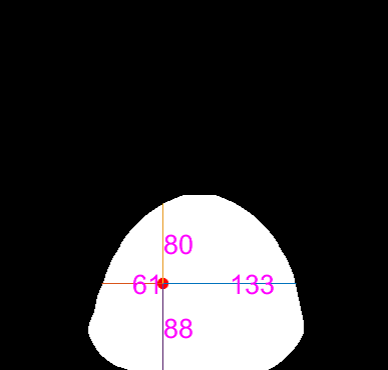}\label{fig:p4_5_img}}
\\
\subfloat[8\_5.tif]{\includegraphics[height=1.1in]{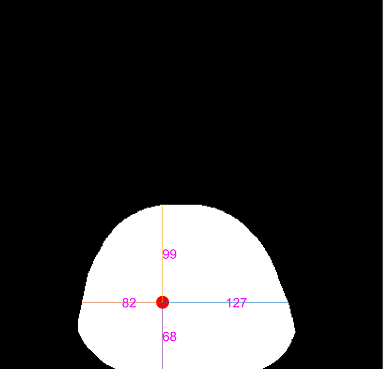}\label{fig:p8_5_img}} 
\thinspace\thinspace
\subfloat[28\_7.tif]{\includegraphics[height=1.1in]{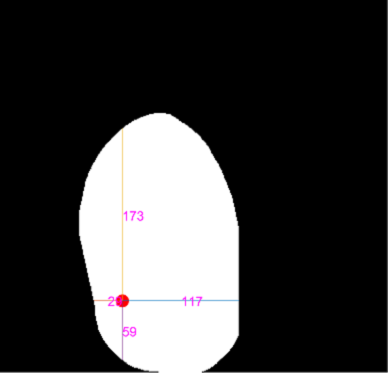}\label{fig:p28_7_img}}
\\
\subfloat[28\_8.tif]{\includegraphics[height=1.1in]{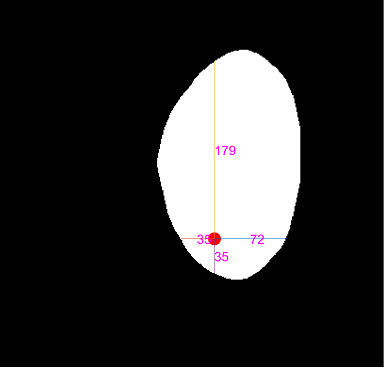}\label{fig:p28_8_img}}
\\
\caption{Outputs from proposed algorithm. Except for 1\_1.tif, rest are partial images for threshold less than \(P_T = 0.6\)}
\end{figure}

For equations (\ref{eq:LC}) to (\ref{eq:DC}) the coordinates of the core point are \((C_X,C_Y)\) while the coordinates for black pixels are \((B_X,B_Y)\)
Equations (\ref{eq:LCN}) to (\ref{eq:DCN}) are normalized values. 
\\\\
Consider equation(\ref{eq:RC}), starting from \(C_X \), white pixels are counted till a black pixel is encountered at \(SegmentationMask(B_{X2},Y)\) or if we encounter the image end, we stop the pixel count. The same analogy is followed for rest of the equations.
\\
To differentiate partial image from non-partials, it can be observed that in partial fingerprint images, the pixel count in any one of the four counts is less than the other three,see Fig.(\ref{fig:p4_5_img}) to Fig.(\ref{fig:p28_8_img}). While in full fingerprint images, this is not the case, see Fig.(\ref{fig:full_img}). 
\\Using the normalized values, we select the minimum pixel count from equation (\ref{eq:MINV})
\begin{equation}\label{eq:MINV} 
Min_{value} = max(L_{cn},R_{cn},U_{cn},D_{cn})
\end{equation}
Applying a threshold \(T<=0.6\),a differentiation of  partial from non-partial fingerprints is achieved.

\section{Results} \label{Results}
The proposed algorithm applied to the local dataset as well as FVC-2002 DB\_1a dataset. There are total 800 images in the data set out of which 62 are partial images. Rest of of them are non-partial images.
\\
Table \ref{ConfusionMatrix} shows the confusion matrix that describes the performance of the proposed algorithm. 
The terms related to the confusion matrix are as follows;

\begin{itemize}
  \item {\textbf{True Positives (TP)}}: Number of partials correctly classified.
  \item {\textbf{True Negatives (TN)}}: Number of non-partials correctly classified.
  \item {\textbf{True Negatives (TN)}}: Number of non-partials correctly classified.
  \item {\textbf{False Negatives (TN)}}: Number of non-partials incorrectly classified.
\end{itemize}

\begin{table}[h]
\begin{center}
\begin{tabular}{ |c|c|c| } 
 \hline
 800 images & Predicted Partial & Predicted Non-Partial \\ 
 \hline
 True Partial & TN = 36 & FP = 18 \\ 
 \hline
 True Non-Partial & FN = 28 & TP = 718 \\ 
 \hline
\end{tabular}
\end{center}
\caption{Confusion matrix for FVC 2002 DB1\_a tests. }
\label{ConfusionMatrix}
\end{table}

Taking into account the above confusion matrix terms, the performance of the algorithm is defined by the following equations.

\begin{equation}\label{eq:Sensitivity} 
Sensitivity = TP/(TP+FN)
\end{equation}

\begin{equation}\label{eq:specificity} 
Specificity = TN/(TN+FN)
\end{equation}

\begin{equation}\label{eq:accuracy} 
Accuracy = (TP+TN)/(TN+FN+TP+FN)
\end{equation}

Sensitivity (also called the true positive rate, the recall, or probability of detection) measures the proportion of positives that are correctly identified.
\\
Specificity (also called the true negative rate) measures the proportion of negatives that are correctly identified.
\\ 
Accuracy is the proportion of the total number of predictions that were correct.
\\

It will interesting to note that a human observer was used to classify partial and non-partial images in order to perform the tests.
\\
\begin{table}[h]
\begin{center}
 \begin{tabular}{|c | c | c | c|} 
 \hline
 Database & Sensitivity & Specificity & Accuracy \\ [0.5ex] 
 \hline
FVC 2002 DB\_1A & 66.6\% & 96.2\% & 94.2\% \\ 
 \hline
\end{tabular}
\end{center}
\caption{Perforance table for proposed method. }
\end{table}

\section{Conclusion} \label{Conclusion}
In this paper we have presented a very simple yet accurate algorithm to detect partial images. The power of the algorithm can be seen from performance table. The algorithm itself is computationally in-expensive and can be used for real time acquisitions as a single image can be processed under 0.1 seconds. The environment under which testing was evaluated was done using Matlab (2015) for Windows 10 using a single thread process on a Core-i7 laptop machine. 

\section*{Acknowledgment}

This research is supported by the Technical Directorate of National Database and Registration Authority's (NADRA) \cite{nadra} Fund for Research And Development in advanced computing.

\end{document}